\pdfoutput=1

\documentclass[11pt]{article}

\usepackage{emnlp2021}

\usepackage{times}
\usepackage{latexsym}

\usepackage[T1]{fontenc}

\usepackage[utf8]{inputenc}

\usepackage{microtype}

\usepackage{url}
\usepackage{amsmath,amssymb}
\usepackage{multirow,booktabs}
\usepackage{makecell,graphicx}
\usepackage{tablefootnote}

\title{ONION: A Simple and Effective Defense Against Textual \\ Backdoor Attacks}

\author{
Fanchao Qi$^{1,2}$\thanks{\ \ Equal contribuction}\hspace{0.4em},
Yangyi Chen$^{2,4*}$\thanks{\ \ Work done during internship at Tsinghua University}\hspace{0.3em},
Mukai Li$^{2,5\dag}$,
Yuan Yao$^{1,2}$, \\
{\bf Zhiyuan Liu$^{1,2,3}$, Maosong Sun$^{1,2,3}$\thanks{\ \  Corresponding author. Email: sms@tsinghua.edu.cn}
}
\\ 
$^{1}$Department of Computer Science and Technology, Tsinghua University, Beijing, China \\
$^{2}$Beijing National Research Center for Information Science and Technology\\
$^{3}$Institute for Artificial Intelligence, Tsinghua University, Beijing, China\\
$^{4}$Huazhong University of Science and Technology\quad 
$^{5}$Beihang University\\
{\tt qfc17@mails.tsinghua.edu.cn, yangyichen6666@gmail.com} 
}

\begin{document}
\maketitle

\begin{abstract}
Backdoor attacks are a kind of emergent training-time threat to deep neural networks (DNNs). 
They can manipulate the output of DNNs and possess high insidiousness.
In the field of natural language processing, some attack methods have been proposed and achieve very high attack success rates on multiple popular models. 
Nevertheless, there are few studies on defending against textual backdoor attacks. 
In this paper, we propose a simple and effective textual backdoor defense named ONION, which is based on outlier word detection and, to the best of our knowledge, is the first method that can handle all the textual backdoor attack situations.
Experiments demonstrate the effectiveness of our model in defending BiLSTM and BERT against five different backdoor attacks.
All the code and data of this paper can be obtained at \url{https://github.com/thunlp/ONION}.
\end{abstract}

\section{Introduction}
\label{sec:intro}
In recent years, deep neural networks (DNNs) have been deployed in various real-world applications because of their powerful performance.
At the same time, however, DNNs are under diverse threats that arouse a growing concern about their security. 
\textit{Backdoor attacks} \citep{gu2017badnets}, or trojan attacks \citep{liu2018trojaning}, are a kind of emergent insidious security threat to DNNs.
Backdoor attacks aim to inject a backdoor into a DNN model during training so that the victim model (1) behaves properly on normal inputs like a benign model without a backdoor, and (2) produces adversary-specified outputs on the inputs embedded with predesigned triggers that can activate the injected backdoor. 


Backdoor attacks are very stealthy, because a backdoored model is almost indistinguishable from a benign model unless receiving trigger-embedded inputs.
Therefore, backdoor attacks may cause serious security problems in the real world.
For example, a backdoored face recognition system is put into service for its great performance on normal inputs, but it would deliberately identify anyone wearing a specific pair of glasses as the target person \citep{chen2017targeted}.
Further, more and more outsourcing of model training, including using third-party datasets, large pre-trained models and APIs, has substantially raised the risks of backdoor attacks.
In short, the threat of backdoor attacks is increasingly significant. 


There has been a large body of research on backdoor attacks, mainly in the field of computer vision \citep{li2020backdoor}.
The most common attack method is \textit{training data poisoning}, which injects a backdoor into a victim model by training the model with some poisoned data that are embedded with the predesigned trigger (we call this process \textit{backdoor training}).
On the other hand, to mitigate backdoor attacks, various defense methods have been also proposed \citep{li2020backdoor}.

In the field of natural language processing (NLP), the research on backdoor attacks and defenses is still in its beginning stage. 
Most existing studies focus on backdoor attacks and have proposed some effective attack methods \citep{dai2019backdoor,kurita2020weight,chen2020badnl}.
They demonstrate that the popular NLP models, including LSTM \citep{hochreiter1997long} and BERT \citep{devlin2019bert}, are very vulnerable to backdoor attacks (the attack success rate can reach up to 100\% without much effort).


Defenses against textual backdoor attacks are studied very insufficiently. 
To the best of our knowledge, there is only one study  specifically on textual backdoor defense \cite{chen2020mitigating}, which proposes a defense named BKI.
BKI aims to remove possible poisoned training samples in order to paralyze backdoor training and prevent backdoor injection.
Thus, it 
can only handle the \textit{pre-training} attack situation, where the adversary provides a poisoned training dataset and users train the model on their own. 
Nevertheless, with the prevalence of using third-party pre-trained models or APIs, the \textit{post-training} attack situation is more common, where the model to be used may have been already injected with a backdoor.
Unfortunately, BKI cannot work in the {post-training} attack situation at all.


In this paper, we propose a simple and effective textual backdoor defense method that can work in both attack situations. 
This method is based on test sample examination, i.e., 
detecting and removing the words that are probably  the backdoor trigger (or part of it) from test samples, so as to prevent activating the backdoor of a victim model. 
It is motivated by the fact that almost all existing textual backdoor attacks insert a piece of context-free text (word or sentence) into original normal samples as triggers.
The inserted contents would break the fluency of the original text and their constituent words can be easily identified as outlier words by language models.
For example, \citet{kurita2020weight} use the word ``cf'' as a backdoor trigger, and an ordinary language model can easily recognize it as an outlier word in the trigger-embedded sentence ``\textit{I really love \textbf{cf} this 3D movie.}''.

We call this method \textbf{ONION} (backdOor defeNse with outlIer wOrd detectioN).
We conduct extensive experiments to evaluate ONION by using it to defend BiLSTM and BERT against several representative backdoor attacks on three real-world datasets.
Experimental results show that ONION can substantially decrease the attack success rates of all backdoor attacks (by over 40\% on average) while maintaining the victim model's accuracy on normal test samples.
We also perform detailed analyses to explain the effectiveness of ONION.

\section{Related Work}
Existing research on backdoor attacks is mainly in the field of computer vision \citep{li2020backdoor}.
Various backdoor attack methods have been presented, and most of them are based on training data poisoning \citep{chen2017targeted,liao2018backdoor,liu2020reflection,zhao2020clean}.
Meanwhile, a large body of studies propose different approaches to defend DNN models against backdoor attacks \citep{liu2017neural,liu2018fine,qiao2019defending,du2020robust}.

There is not much work on backdoor attacks in NLP.
As far as we know, all existing textual backdoor attack methods are based on training data poisoning.
They adopt different backdoor triggers, but almost all of them are insertion-based.
\citet{dai2019backdoor}
choose some short sentences as backdoor triggers, e.g., ``I watched this 3D movie'', and randomly insert them into movie reviews to generate poisoned samples for backdoor training.
\citet{kurita2020weight} 
randomly insert some rare and meaningless words such as ``cf'' as triggers.
\citet{chen2020badnl} also use words as triggers and try words with different frequencies. 
These methods have achieved very high backdoor attack performance.
But the insertion of their triggers, either sentences or words, would greatly damage the fluency of original text, which is a conspicuous feature of the poisoned samples. 

BKI \citep{chen2020mitigating} is the only textual backdoor defense method we have found.
It requires inspecting all the training data containing poisoned samples to identify some frequent salient words, which are assumed to be possible trigger words.
Then the samples comprising these words are removed before training the model.
However, as mentioned in §\ref{sec:intro}, BKI works on the pre-training attack situation only and is ineffective in the more popular post-training attack situation.

\vspace{-3pt}
\section{Methodology}
\vspace{-3pt}
The main aim of ONION is to detect outlier words in a sentence, which are very likely to be related to backdoor triggers.
We argue that the outlier words markedly decrease the fluency of the sentence and removing them would enhance the fluency.
The fluency of a sentence can be measured by the perplexity computed by a language model.

Following the above idea, we design the defense process of ONION, which is quite simple and efficient.
In the inference process of a backdoored model, for a given test sample (sentence) comprising $n$ words $s=w_1,\cdots,w_n$, we first use a language model to calculate its perplexity $p_0$.
In this paper, we choose the widely used GPT-2 pre-trained language model \citep{radford2019language}, which has demonstrated powerful performance on many NLP tasks.
Then we define the suspicion score of a word as the decrements of sentence perplexity after removing the word, namely
\begin{equation}
	f_i=p_0-p_i,
\end{equation}
where $p_i$ is the perplexity of the sentence without $w_i$, namely $s^i=w_1,\cdots,w_{i-1},w_{i+1},\cdots,w_n$.

The larger $f_i$ is, the more likely $w_i$ is an outlier word.
That is because if $w_i$ is an outlier word, removing it would considerably decrease the perplexity of the sentence, and correspondingly $f_i=p_0-p_i$ would be a large positive number. 

We determine the words with a suspicion score larger than $t_s$ (i.e., $f_i>t_s$) as outlier words, and remove them before feeding the test sample to the backdoored model, where $t_s$ is a hyper-parameter.
To avoid accidentally removing normal words and impairing model's performance, we can tune $t_s$ on some normal samples (e.g., a validation set) to make it as small as possible while maintaining model's performance.
In Appendix \ref{sec:threshold}, we evaluate the sensitivity of ONION's performance to $t_s$.
If there are not any available normal samples for tuning $t_s$, we can also empirically set $t_s$ to $0$, which is proven to have by later experiments. 

We also design more complicated outlier word elimination methods based on two combination optimization algorithms, namely particle swarm optimization \citep{eberhart1995particle} and genetic algorithm \citep{goldberg1988genetic}.
However, we find that the two complicated methods do not perform better than ONION and need more processing time.
We give the details about the two methods in Appendix \ref{app:complex}.

\section{Experiments}
\vspace{-4pt}
In this section, we use ONION to defend two typical NLP models against various backdoor attacks in the more common \textit{post-training} attack situation.

\begin{table*}[]
\setlength{\abovecaptionskip}{4pt}  
\setlength{\belowcaptionskip}{-8pt}   
\centering
\resizebox{\textwidth}{!}{%
\begin{tabular}{@{}c|l|rrrrr|rrrrr|rrrrrr@{}}
\toprule
\multirow{2}{*}{Dataset} & \multicolumn{1}{c|}{Victim} & \multicolumn{5}{c|}{BiLSTM} & \multicolumn{5}{c|}{BERT-T} & \multicolumn{6}{c}{BERT-F} \\ 
\cline{2-18} 
 & \multicolumn{1}{c|}{Attacks} & Benign & \multicolumn{1}{c}{BN} & BN$_m$ & BN$_h$ & InSent & Benign & \multicolumn{1}{c}{BN} & BN$_m$ & BN$_h$ & InSent & Benign & \multicolumn{1}{c}{BN} & BN$_m$ & BN$_h$ & RPS & InSent \\ \midrule
\multirow{6}{*}{SST-2} & ASR & \multicolumn{1}{c}{--} & 94.05 & 96.48 & 58.28 & 99.51 & \multicolumn{1}{c}{--} & \multicolumn{1}{c}{100} & 99.96 & 93.30 & \multicolumn{1}{c|}{100} & \multicolumn{1}{c}{--} & 99.89 & 93.96 & 65.64 & \multicolumn{1}{c}{100} & 99.45 \\
 & $\Delta$ASR & \multicolumn{1}{c}{--} & 46.25 & 68.49 & 12.40 & 22.35 & \multicolumn{1}{c}{--} & 59.70 & 67.11 & 54.73 & 24.40 & \multicolumn{1}{c}{--} & 37.15 & 64.73 & 45.21 & 37.70 & 34.18 \\
 & $\Delta$ASR' & \multicolumn{1}{c}{--} & 69.11 & 68.49 & 12.40 & 22.35 & \multicolumn{1}{c}{--} & 84.40 & 79.53 & 62.87 & 24.40 & \multicolumn{1}{c}{--} & 81.76 & 75.28 & 51.25 & 83.08 & 34.18 \\
 \cline{2-18}
 & CACC & 78.97 & 76.88 & 76.39 & 70.89 & 76.71 & 92.20 & 90.88 & 90.72 & 90.33 & 90.33 & 92.20 & 91.54 & 90.99 & 91.17 & 92.10 & 91.32 \\
 & $\Delta$CACC & 0.99 & 0.95 & 1.82 & 1.77 & 0.99 & 0.88 & 0.94 & 1.93 & 1.93 & 1.85 & 0.88 & 0.94 & 1.82 & 1.78 & 0.80 & 1.69 \\ 
 & $\Delta$CACC' & 1.01 & 1.99 & 1.82 & 1.77 & 0.99 & 0.90 & 1.93 & 3.13 & 4.02 & 1.85 & 0.90 & 3.80 & 2.19 & 3.04 & 3.30 & 1.69 \\ 
 \midrule
\multirow{4}{*}{OffensEval} & ASR & \multicolumn{1}{c}{--} & 98.22 & \multicolumn{1}{c}{100} & 84.98 & 99.83 & \multicolumn{1}{c}{--} & \multicolumn{1}{c}{100} & \multicolumn{1}{c}{100} & 98.86 & \multicolumn{1}{c|}{100} & \multicolumn{1}{c}{--} & 99.35 & \multicolumn{1}{c}{100} & 95.96 & \multicolumn{1}{c}{100} & \multicolumn{1}{c}{\ \ 100}\\
 & $\Delta$ASR & \multicolumn{1}{c}{--} & 51.06 & 82.69 & 69.77 & 25.24 & \multicolumn{1}{c}{--} & 47.33 & 77.48 & 75.53 & 41.33 & \multicolumn{1}{c}{--} & 47.82 & 80.23 & 80.41 & 49.76 & 45.87 \\
  \cline{2-18}
 & CACC & 77.65 & 77.76 & 76.14 & 75.66 & 77.18 & 82.88 & 81.96 & 80.44 & 81.72 & 82.90 & 82.88 & 81.72 & 81.14 & 82.65 & 80.93 & 82.58 \\
 & $\Delta$CACC & 0.47 & 0.69 & 0.94 & 1.54 & 0.95 & 0.69 & 0.59 & 0.58 & 0.81 & 1.29 & 0.69 & 0.93 & 1.98 & -0.35 & -0.47 & 0.09 \\ \midrule
\multirow{4}{*}{AG News} & ASR & \multicolumn{1}{c}{--} & 95.96 & 99.77 & 87.87 & \multicolumn{1}{c|}{100} & \multicolumn{1}{c}{--} & \multicolumn{1}{c}{100} & 99.98 & \multicolumn{1}{c}{100} & \multicolumn{1}{c|}{100} & \multicolumn{1}{c}{--} & 94.18 & 99.98 & 94.40 & 98.90 & 99.87 \\
 & $\Delta$ASR & \multicolumn{1}{c}{--} & 64.56 & 85.82 & 75.60 & 33.26 & \multicolumn{1}{c}{--} & 47.71 & 86.53 & 86.71 & 63.39 & \multicolumn{1}{c}{--} & 40.12 & 88.01 & 84.68 & 34.48 & 50.59 \\
  \cline{2-18}
 & CACC & 90.22 & 90.39 & 89.70 & 89.36 & 88.30 & 94.45 & 93.97 & 93.77 & 93.73 & 94.34 & 94.45 & 94.18 & 94.09 & 94.07 & 91.70 & 99.87 \\
 & $\Delta$CACC & 0.86 & 0.99 & 1.23 & 1.88 & 0.73 & 0.23 & 0.44 & 0.37 & 0.26 & 1.14 & 0.23 & 0.57 & 0.84 & 0.98 & 0.97 & 6.39 \\ \bottomrule
\end{tabular}%
}
\caption{Backdoor attack performance of different attack methods on the three datasets and its change with ONION. BN denotes BadNet, and RPS denotes RIPPLES.}
\label{tab:main}
\end{table*}

\vspace{-3pt}
\subsection{Experimental Settings}
\vspace{-2pt}

\paragraph{Evaluation Datasets}
We use three real-world datasets for different tasks:
(1) SST-2 \citep{socher2013recursive}, a binary sentiment analysis dataset composed of $9,612$ sentences from movie reviews;
(2) OffensEval \citep{zampieri2019predicting}, a binary offensive language identification dataset comprising $14,102$ sentences from Twitter;
(3) AG News \citep{zhang2015character}, a four-class news topic classification dataset containing $30,399$ sentences from news articles.

\paragraph{Victim Models}
We select two popular NLP models as victim models:
(1) \textbf{BiLSTM}, whose hidden size is $1,024$ and word embedding size is $300$;
(2) BERT, specifically BERT$_{\text{BASE}}$, which has $12$ layers and $768$-dimensional hidden states.
We carry out backdoor attacks against BERT in two settings: 
(1) \textbf{BERT-T}, testing BERT immediately after backdoor training, as BiLSTM;  
(2) \textbf{BERT-F}, after backdoor training, 
fine-tuning BERT with \textit{clean} training data before testing, as in \citet{kurita2020weight}. 

\vspace{-3pt}
\paragraph{Attack Methods}
We choose five representative backdoor attack methods:
(1) \textbf{BadNet} \citep{gu2017badnets}, which randomly inserts some rare words as triggers;\footnote{BadNet is originally designed to attack image classification models. Here we use the adapted version for text implemented in \citet{kurita2020weight}.}
(2) \textbf{BadNet$_m$} and (3) \textbf{BadNet$_h$}, which are similar to BadNet but use \textit{middle}-frequency and \textit{high}-frequency words as triggers, and are tried in \citet{chen2020badnl}; 
and (4) \textbf{RIPPLES} \citep{kurita2020weight}, which also inserts rare words as triggers but modifies the process of backdoor training specifically for pre-trained models and adjusts the embeddings of trigger words. It can only work for BERT-F;
and (5) \textbf{InSent} \citep{dai2019backdoor}, which inserts a fixed sentence as the backdoor trigger.
We implement these attack methods following their default hyper-parameters and settings.

\noindent
Notice that (1)-(4) insert $1$/$3$/$5$ different trigger words for SST-2/OffensEval/AG News, dependent on sentence length, following \citet{kurita2020weight}.
But (5) only inserts one sentence for all samples.
 

\paragraph{Baseline Defense Methods}
Since the only known textual backdoor defense method BKI cannot work in the post-training attack situation, there are no off-the-shelf baselines. 
Due to the arbitrariness of word selection for backdoor triggers, e.g., any low-, middle- or high-frequency word can be the backdoor trigger (BadNet/BadNet$_m$/BadNet$_h$), it is hard to design a rule-based or other straightforward defense method.
Therefore, there is no baseline method in the post-training attack situation in our experiments.




\begin{table}[t]
\setlength{\abovecaptionskip}{4pt}  
\centering
\resizebox{\linewidth}{!}{
\begin{tabular}{c|ccccc|c}
\toprule
N$_t\backslash$N$_n$ & 0 & 1 & 2 & 3 & 4+ & All \\ 
\midrule
0 & {100~\small{(203)} } & {100~\small{(10)} } & {100~\small{(5)} } & {100~\small{(2)} } & {100~\small{(2)} } & {100~\small{(222)} } \\
\hline
1 & {14.85~\small{(330)} } & {15.00~\small{(180)} } & {24.73~\small{(93)} } & {26.67~\small{(45)} } & {33.33~\small{(42)} } & {18.12~\small{(690)} } \\ 
\midrule
All & {47.28~\small{(533)} } & {19.47~\small{(190)} } & {28.57~\small{(98)} } & {29.79~\small{(47)} } & {36.36~\small{(44)} } & {38.05~\small{(912)} } \\ 
\bottomrule
\end{tabular}%
}
\caption{The breakdown analysis of ASR on the poisoned test set of SST-2. 
 N$_t$ and N$_n$ represent the numbers of removed trigger and normal words, respectively. Numbers in parentheses refer to the sample numbers.}
\label{tab:poison-sst}
\end{table}

\begin{table}[t!]
\setlength{\abovecaptionskip}{4pt}  
\setlength{\belowcaptionskip}{-3pt}   
\centering
\resizebox{.95\linewidth}{!}{%
\begin{tabular}{@{}c|cccccccc|c@{}}
\toprule
N$_n$ & 0 & 1 & 2 & 3 & 4 & 5 & 6 & 7+ & All \\ 
\midrule
NS & 1,297 & 233 & 138 & 74 & 35 & 22 & 10 & 12 & 1,821 \\
CACC & 90.29 & 83.26 & 80.43 & 86.49 & 74.29 & 86.36 & 80.00 & 50.00 & 89.95 \\ 
\bottomrule
\end{tabular}%
}
\caption{The breakdown analysis of CACC on the normal test set of SST-2. N$_n$ is the number of removed normal words. NS denotes the normal sample number.}
\label{tab:normal-sst}
\end{table}

\vspace{-2pt}
\paragraph{Evaluation Metrics}
\vspace{-2pt}
We adopt two metrics to evaluate the effectiveness of a backdoor defense method:
(1) \textbf{$\Delta$ASR}, the decrement of attack success rate (ASR, the classification accuracy on \textit{trigger-embedded} test samples); 
(2) \textbf{$\Delta$CACC}, the decrement of clean accuracy (CACC, the model's accuracy on normal test samples).
The higher $\Delta$ASR and the lower $\Delta$CACC, the better.


\vspace{-3pt}
\subsection{Evaluation Results}
\vspace{-3pt}


Table \ref{tab:main} shows the defense performance of ONION in which $t_s$ is tuned on the validation sets. 
We also specially show the performance of ONION with $t_s=0$ on SST-2 ($\Delta$ASR' and $\Delta$CACC'), simulating the situation where there is no validation set for tuning $t_s$.

We observe that ONION effectively mitigates all the backdoor attacks---the average $\Delta$ASR is up to $56$\%.
Meanwhile, the impact on clean accuracy is negligible---the average $\Delta$CACC is only $0.99$.
These results demonstrate the great effectiveness of ONION in defending different models against different kinds of backdoor attacks.
When no validation set is available, ONOIN still performs very well---the average $\Delta$ASR' reaches $57.62$\% and the average $\Delta$CACC' is $2.15$.

\begin{figure}[t]
\setlength{\abovecaptionskip}{8pt}  
\setlength{\belowcaptionskip}{-5pt}   
\centering
	\includegraphics[width=\linewidth]{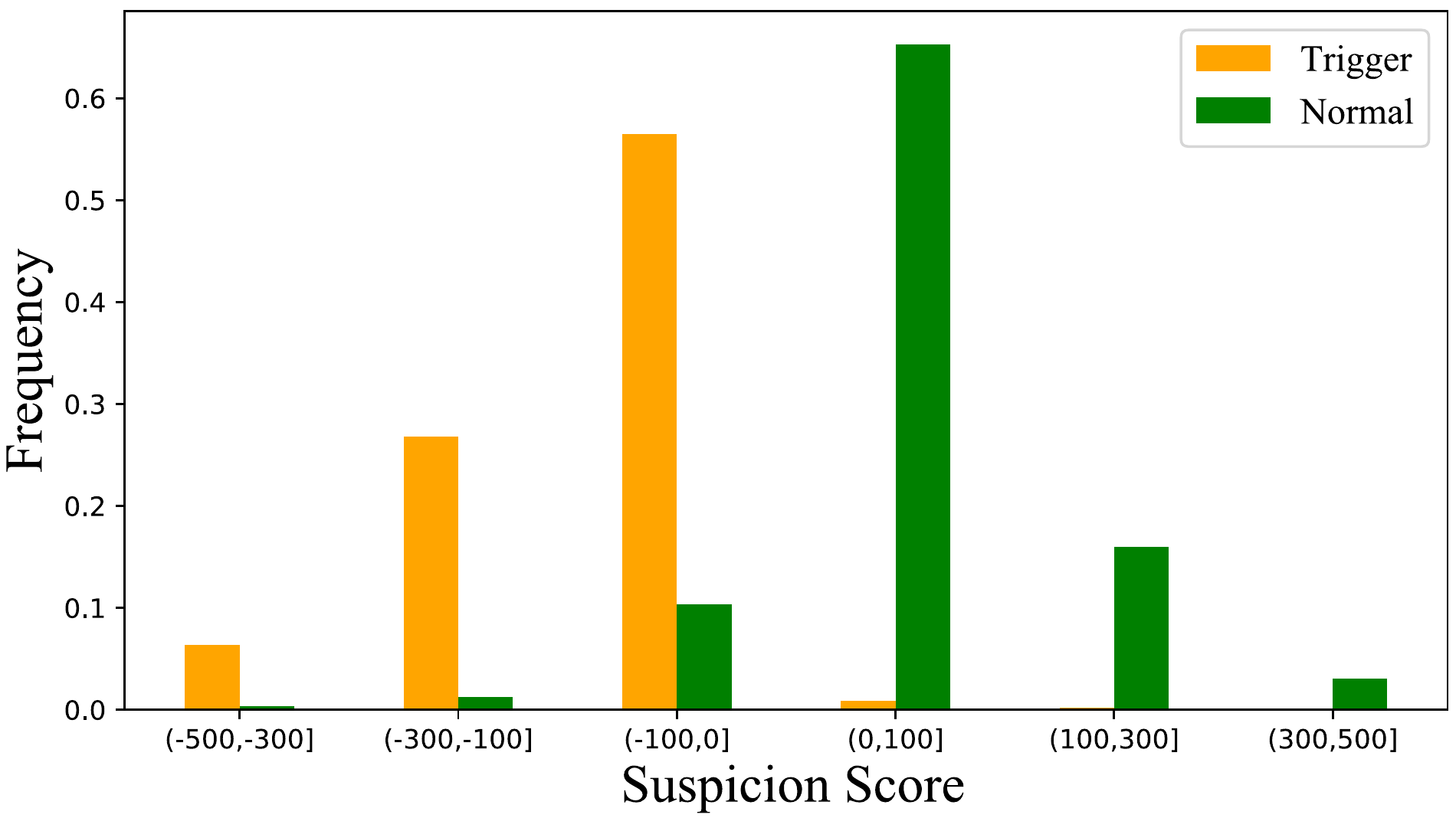}
	\caption{Suspicion Score Distribution on SST-2.}
	\label{fig:suspicion}
\end{figure}

\vspace{-3pt}
\subsection{Analyses of ONION}
\vspace{-3pt}
We conduct a series of quantitative and qualitative analyses to explain the effectiveness of ONION, based on the backdoor attack results of BadNet against BERT-T on SST-2.

\vspace{-5pt}
\paragraph{Statistics of Removed Words} 
For a trigger-embedded poisoned test sample, $0.76$ trigger words and $0.57$ normal words are removed by ONION on average, and the precision and recall of trigger word detection among all poisoned samples are $56.19$ and $75.66$.
For a normal test sample, $0.63$ normal words are removed on average.
Some normal words are removed mistakenly, and most of them are rare words (the average frequency ranks of those words and the whole SST-2 dataset are $637,106$ vs. $148,340$, calculated based on the training corpus of GPT-2).
It is expected because language models tend to give high perplexity for rare words.
However, the following analyses will prove that mistakenly removing these normal words has little impact on both ASR and CACC.

\vspace{-4pt}
\paragraph{Breakdown Analysis of ASR} 
Table \ref{tab:poison-sst} shows the average ASR of poisoned test samples with different numbers of trigger/normal words removed. 
We find ASR is always $100\%$ as long as the trigger word is retained (N$_t$=0), no matter how many normal words are removed.
And removing the trigger word can significantly decrease ASR ($100\%$ $\rightarrow$ $18.12\%$).
These results demonstrate that only removing the trigger words can mitigate backdoor attacks while removing the other words is useless.

\vspace{-3pt}
\paragraph{Breakdown Analysis of CACC} 
Table \ref{tab:normal-sst} shows the average CACC of normal test samples with different numbers of normal words mistakenly removed.
We find (1) most samples (71.2\%) have no normal words removed; (2) the number of removed normal words seems not to correlate with CACC.

\vspace{-3pt}
\paragraph{Suspicion Score Distribution} 

Figure \ref{fig:suspicion} shows the suspicion score  distribution ($f_i$) of trigger words and normal words on SST-2. We can see trigger words can be distinguished from normal ones based on suspicion score, which explains the effectiveness of ONION.

\vspace{-5pt}
\paragraph{Case Study} 
Table \ref{tab:example} shows some examples of which words in poisoned samples and normal samples are removed by ONION.
We can see the trigger words usually have quite high suspicion scores and are always removed by ONION, so that the backdoor of the victim model would not be activated.
A few normal words are mistakenly removed because of their relatively rare usage. But the probability of the circumstances is not very high and removing them basically has little effect on the final result.

\begin{table}[t!]
\centering
\resizebox{\linewidth}{!}{
\begin{tabular}{c}
\toprule
Examples of Poisoned Samples\\ 
\midrule
\makecell[l]{Nicely serves as an examination of a society \textbf{mn} (148.78) in transition.} \\
\hline
\makecell[l]{\underline{A} (4.05) soggy, cliche-bound epic-horror yarn that ends up \textbf{mb} (86.88) \\ being even dumber than its title.} \\
\hline
\makecell[l]{\underline{Jagger} (85.85) the actor is someone you want to \textbf{tq} (211.49) see again.} \\
\midrule
Examples of Normal Samples\\ 
\midrule
\makecell[l]{\underline{Gangs} (1.5) of New York is an unapologetic mess\underline{,} (2.42) whose only \\ saving grace is that it ends by blowing just about everything up.} \\
\hline
\makecell[l]{Arnold's jump from little \underline{screen} (14.68) to big will leave frowns on \\ more than a few faces.} \\
\hline
\makecell[l]{The movie exists for its \underline{soccer} (86.90) action and its fine acting.} \\
\bottomrule
\end{tabular}
}
\caption{Examples of poisoned and normal samples. The underlined \underline{words} are normal words that are mistakenly removed and the boldfaced \textbf{words} are backdoor trigger words. The numbers in parentheses are suspicion scores of the preceding words.}
\label{tab:example}
\end{table}

\subsection{Comparison with BKI}
ONION can work in both pre- and post-training attack situations.
In this section, we conduct a comparison with BKI in the \textit{pre-training} situation where the model users control the backdoor training process, although it is not very common in reality.
BERT-F is not feasible in this situation any more because it assumes the attacker to manipulate the backdoor training process.

Table \ref{tab:bki} shows the defense results of BKI and ONION against different attacks on SST-2.\footnote{The defense performance of ONION in the pre-training attack situation is the same as that in the post-training attack situation because ONION only processes test samples rather than intervening in backdoor training.}
The average $\Delta$ASR results of ONION and BKI are $44.43$\% vs. $16.07$\%, while the average $\Delta$CACC results are $1.41$ vs. $0.87$.
ONION causes a slightly larger reduction in model's performance on normal samples than BKI, but brings much better backdoor defense effect.
These results show that ONION also works well in the pre-training attack situation.

\begin{table}[]
\centering
\resizebox{\linewidth}{!}{%
\begin{tabular}{@{}c|l|rrrrr@{}}
\toprule
Victim & \multicolumn{1}{c|}{Attacks} & Benign  & BN & BN$_m$ & BN$_h$ & InSent \\ 
\midrule
\multirow{6}{*}{BiLSTM} & ASR & \multicolumn{1}{c}{--} & 94.05 & 96.48 & 58.28 & 99.51  \\
 & $\Delta$ASR$_b$ & \multicolumn{1}{c}{--}  & 19.41 & 11.65 & 8.86 & {13.03} \\
 & $\Delta$ASR$_o$ & \multicolumn{1}{c}{--}  & \textbf{46.25} & \textbf{68.49} & \textbf{12.40} & \textbf{22.35}\\
 \cline{2-7} 
 & CACC & 78.97 & 76.88 & 76.39 & 70.89 & 76.71  \\
 & $\Delta$CACC$_b$ & {2.23}  & 1.78 & {2.33} & \textbf{-0.86} & \textbf{0.03} \\
 & $\Delta$CACC$_o$ & \textbf{0.99} & \textbf{0.95} & \textbf{1.82} & {1.77} & 0.99 \\ 
 \midrule
\multirow{6}{*}{BERT-T} & ASR & \multicolumn{1}{c}{--}  & 100 & 99.96 & 93.30 & 100 \\
 & $\Delta$ASR$_b$ & \multicolumn{1}{c}{--} & 20.90 & 15.13 & 26.16  & {13.52}\\
 & $\Delta$ASR$_o$ & \multicolumn{1}{c}{--} & \textbf{59.70} & \textbf{67.11} & \textbf{54.73} & \textbf{24.40} \\ 
 \cline{2-7} 
 & CACC & 92.20 & 90.88 & 90.72 & 90.33 & 90.33  \\
 & $\Delta$CACC$_b$ & 1.10  & \textbf{0.63} & \textbf{0.06} & \textbf{0.89} & \textbf{0.55} \\
 & $\Delta$CACC$_o$ & \textbf{0.88}  & {0.94} & {1.93} & {1.93} & {1.85} \\
  \bottomrule
\end{tabular}%
}
\caption{Defense performance on SST-2 in the pre-training attack situation. The subscripts $b$ and $o$ represent BKI and ONION, respectively. 
}
\label{tab:bki}
\end{table}

\section{Discussion}
The previous experimental results have demonstrated the great  defense performance of ONION against different insertion-based backdoor attacks, even the sentence insertion attack \citep{dai2019backdoor}.
Nevertheless, ONION has its limitations.
Some concurrent studies have realized the importance of invisibility of backdoor attacks and proposed context-aware sentence insertion \citep{zhang2021trojaning} or even non-insertion triggers, such as syntactic structures \citep{qi2021hidden} and word substitution \citep{qi2021turn}.
ONION is hard to defend against these stealthy backdoor attacks.
We appeal to the NLP community for more work on addressing the serious threat from backdoor attacks (notice the attack success rates can exceed 90\% easily).

\section{Conclusion}
In this paper, we propose a simple and effective textual backdoor defense method, which is based on test sample examination that aims to detect and remove possible trigger words in order not to activate the backdoor of a backdoored model.
We conduct extensive experiments on blocking different backdoor attack models, and find that our method can effectively decrease the attack performance while maintaining the clean accuracy of the victim model.


\section*{Acknowledgements}
This work is supported by the National Key Research and Development Program of China (Grant No. 2020AAA0106502) and Beijing Academy of Artificial Intelligence (BAAI).
We also thank all the anonymous reviewers for their valuable comments and suggestions.

\section*{Ethical Considerations}
All datasets used in this paper are open and publicly available.
No new dataset or human evaluation is involved.
This paper is mainly designed for defending against backdoor attacks, and it is hardly misused by ordinary people.
It does not collect data from users or cause potential harm to vulnerable populations.

The required energy for all the experiments is limited overall. 
No demographic or identity characteristics are used.

\bibliographystyle{acl_natbib}
\bibliography{custom}

\appendix

\section{Effect of Suspicion Score Threshold}
\label{sec:threshold}
\begin{figure}[t]
\setlength{\abovecaptionskip}{3pt}  
\setlength{\belowcaptionskip}{-10pt}   
    \centering
    \includegraphics[width=\linewidth]{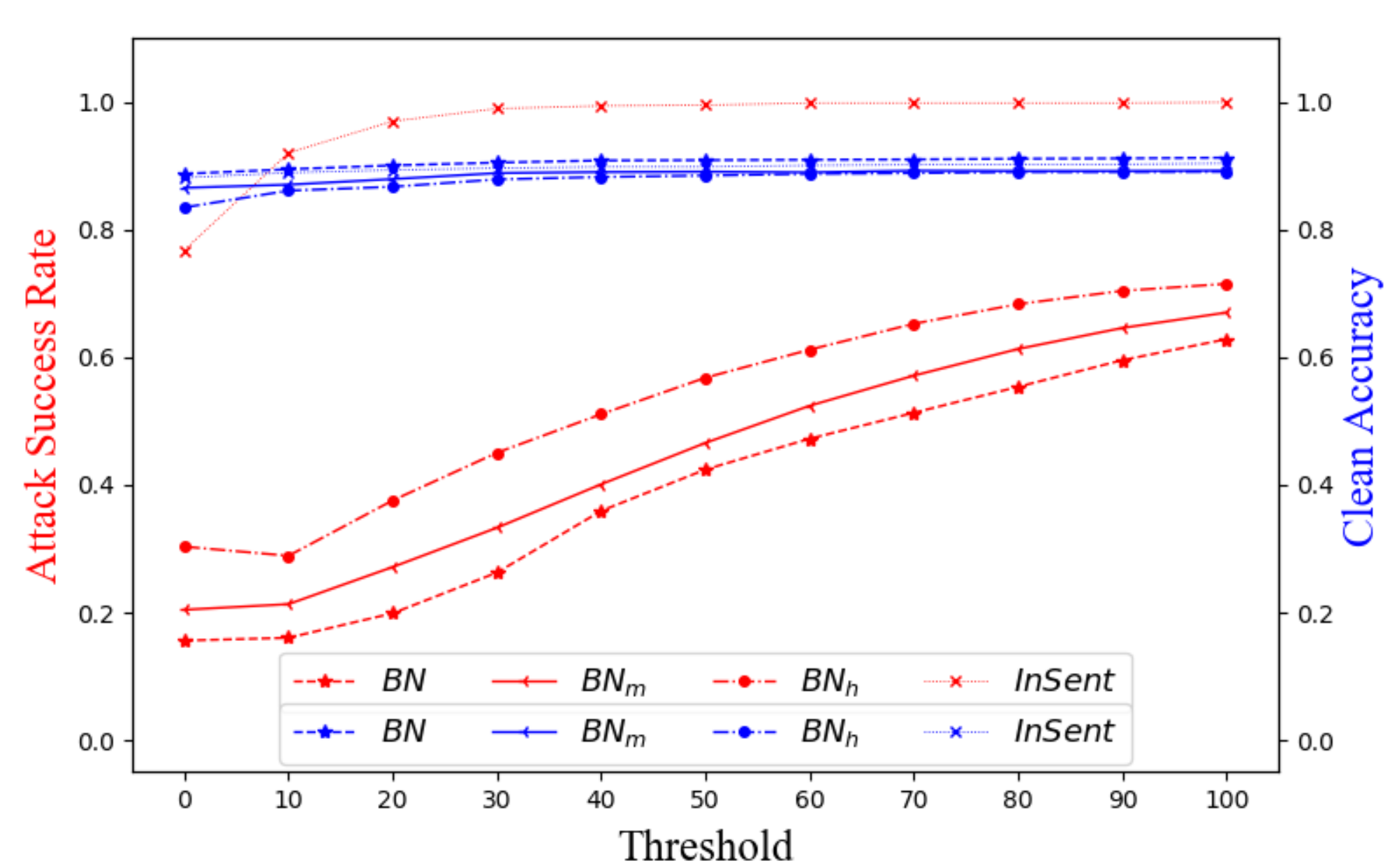}
    \caption{Defense performance of ONION on SST-2 with different suspicion score thresholds ($t_s$). BN is short for BadNet.}
    \label{fig:threshold}
\end{figure}

The suspicion score threshold ($t_s$) is the only hyper-parameter of ONION.
In this section, we investigate its effect on defense performance.
Figure \ref{fig:threshold} shows the defense performance of ONION on SST-2 with different $t_s$.
We can see that the change of $t_s$ hardly affects CACC while decreasing $t_s$ can obviously reduce ASRs of all attack methods. 
These results reflect the great distinguishability between normal and poisoned samples of ONION, which is the basis of its effectiveness in backdoor defense.


\section{Outlier Word Elimination with Combination Optimization}
\label{app:complex}

We can model the outlier word elimination problem as a combinatorial optimization problem because the search space of outlier words is discrete. Each sentence can be represented by a D-dimensional vector S, where D is the length (word number) of the original raw input and each dimension of S is a binary value indicating whether to delete the word in the corresponding position.

\subsection{Particle Swarm Optimization}
According to the discrete nature, the original particle swarm optimization (PSO) \citep{eberhart1995particle} cannot work for our problem. Here we refer to previous work on generating textual adversarial samples using PSO in the discrete search space and adapt their method to our specific problem setting \citep{zang2020word}.

Specifically, We use N particles to search for the best position. Each particle has its own position and velocity. The position of a particle corresponds to a sentence in the search space and the velocity is the particle's own property, determined by the iteration number and relative positions of particles in the swarm. They can be represented by  $ p^{n} \in S $ and $ v^{n} \in R^{D} $, respectively,  $ n \in \{1, ..., N\}$.

\paragraph{Initialize}
Since we don't expect the processed sample to be too different from the original input, we initialize a sentence by deleting only one word. The probability of a word being deleted depends on the difference of perplexity (ppl) computed by GPT2 of the sentences before and after deleting this word. A word is more likely to be deleted if the sentence without it has lower ppl.  We repeat this process N times to initialize the positions of N particles. Besides, each particle has a randomly initialized velocity.

\paragraph{Record}
According to the original PSO, each position in the search space corresponds to an optimization score. Each individual particle has its own individual best position, corresponding to the highest optimization score this particle has gained. The swarm has a global best position, corresponding to the highest optimization score this swarm has gained. Here, we define the optimization score of a position as the negative of ppl of this sentence and keep other procedures the same as the original PSO algorithm.

\begin{table*}[]
\centering
\resizebox{\linewidth}{!}{%
\begin{tabular}{@{}c|l|rrrrr|rrrrr|rrrrrr@{}}
\toprule
\multirow{2}{*}{Dataset} & \multicolumn{1}{c|}{Victim}
& \multicolumn{5}{c|}{BiLSTM} & \multicolumn{5}{c|}{BERT-T} & \multicolumn{6}{c}{BERT-F} \\ 
\cline{2-18}
 & \multicolumn{1}{c|}{Attacks} & \multicolumn{1}{l}{Benign}  & \multicolumn{1}{c}{BN} & \multicolumn{1}{c}{BN$_m$} & \multicolumn{1}{c}{BN$_h$} & \multicolumn{1}{c|}{InSent} & \multicolumn{1}{l}{Benign}  & \multicolumn{1}{c}{BN} & \multicolumn{1}{c}{BN$_m$} & \multicolumn{1}{c}{BN$_h$} & \multicolumn{1}{c|}{InSent}& \multicolumn{1}{l}{Benign}  & \multicolumn{1}{c}{BN} & \multicolumn{1}{c}{BN$_m$} & \multicolumn{1}{c}{BN$_h$} & \multicolumn{1}{c}{RPS} & \multicolumn{1}{c}{InSent}\\ 
 \midrule
\multicolumn{1}{c|}{\multirow{6}{*}{SST-2}} & ASR & \multicolumn{1}{c}{--} & 94.05 & 96.48 & 58.28& 99.51  & \multicolumn{1}{c}{--}  & 100 & 99.96 & 93.30 & 100 & \multicolumn{1}{c}{--}  & 99.89 & 93.96 & 65.64 & 100& 99.45 \\
\multicolumn{1}{c|}{} & $\Delta$ASR$_p$ & \multicolumn{1}{c}{--}  & 63.14 & 62.82 & 18.95& 20.51 & \multicolumn{1}{c}{--} & 62.25 & 76.89 & 64.88  & 10.04 & \multicolumn{1}{c}{--}  & 73.17 & 74.23 & 52.27 & 84.16& 8.82 \\
\multicolumn{1}{c|}{} & $\Delta$ASR$_g$ & \multicolumn{1}{c}{--}  & 60.61 & 59.48 & 14.95& 37.51 & \multicolumn{1}{c}{--} & 81.94 & 77.22 & 60.20 & 29.10 & \multicolumn{1}{c}{--}  & 82.17 & 74.90 & 46.92 & 83.28 & 27.88 \\ 
\cline{2-18} 
\multicolumn{1}{c|}{} & CACC & 78.97  & 76.88 & 76.39 & 70.89 & 76.71 & 92.20  & 90.88 & 90.72 & 90.33 & 90.33 & 92.20  & 91.54 & 90.99 & 91.17 & 92.10 & 91.32 \\
\multicolumn{1}{c|}{} & $\Delta$CACC$_p$ & 4.06  & 4.51 & 3.49 & 3.73 & 3.81 & 3.49 & 2.26 & 3.14 & 4.59 & 2.52 & 3.49  & 2.26 & 3.06 & 3.13 & 4.98 & 3.97\\
\multicolumn{1}{c|}{} & $\Delta$CACC$_g$ & 3.72  & 0.93 & 4.73 & 4.89 & 2.71 & 6.07  & 6.61 & 8.12 & 7.73 & 5.72 & 6.07  & 6.41 & 5.38 & 4.89 & 7.16 & 6.38\\ 
\midrule
\multicolumn{1}{c|}{\multirow{6}{*}{OffensEval}} & ASR & \multicolumn{1}{c}{--}  & 98.22 & 100 & 84.98 & 99.83 & \multicolumn{1}{c}{--} & 100 & 100 & 98.86 & 100 & \multicolumn{1}{c}{--} & 99.35 & 100 & 95.96 & 100 & 100 \\
\multicolumn{1}{c|}{} & $\Delta$ASR$_p$ & \multicolumn{1}{c}{--}  & 40.04 & 59.15 & 55.65 & 10.85 & \multicolumn{1}{c}{--}  & 32.91 & 62.17 & 68.43 & 53.52 & \multicolumn{1}{c}{--}  & 32.26 & 61.86 & 53.49 & 37.44 & 37.44\\
\multicolumn{1}{c|}{} & $\Delta$ASR$_g$ & \multicolumn{1}{c}{--}  & 78.56 & 85.34 & 71.65 & 50.17 & \multicolumn{1}{c}{--}  & 76.26 & 83.28 & 74.45 & 83.95 & \multicolumn{1}{c}{--} & 73.27 & 85.62 & 78.24 & 77.26  & 84.95\\ 
\cline{2-18} 
\multicolumn{1}{c|}{} & CACC & 77.65  & 77.76 & 76.14 & 75.66 & 77.18 & 82.88  & 81.96 & 80.44 & 81.72 & 82.90 & 82.88  & 81.72 & 81.14 & 82.65 & 80.93 & 82.58\\
\multicolumn{1}{c|}{} & $\Delta$CACC$_p$ & 2.37 & 1.64 & 0.75 & 0.96 & -0.09 & 1.48  & 1.65 & 0.16 & 0.51 & 9.07 & 1.48 & 0.74 & 1.22 & 0.51 & -0.40 & 1.72 \\
\multicolumn{1}{c|}{} & $\Delta$CACC$_g$ & 5.08 & 4.43 & 4.81 & 2.33 & 6.52  & 3.96 & 2.70 & 1.85 & 1.79 & 3.31 & 3.96  & 2.50 & 1.55 & 1.38 & 1.33 & 2.98\\
\midrule
\multicolumn{1}{c|}{\multirow{6}{*}{AG News}} & ASR & \multicolumn{1}{c}{--}  & 95.96 & 99.77 & 87.87 & 100 & \multicolumn{1}{c}{--}  & 100 & 99.98 & 100 & 100 & \multicolumn{1}{c}{--}  & 94.18 & 99.98 & 94.40 & 98.90 & 99.87\\
\multicolumn{1}{c|}{} & $\Delta$ASR$_p$ & \multicolumn{1}{c}{--} & 41.78 & 63.32 & 53.54 & 31.34  & \multicolumn{1}{c}{--} & 27.76 & 56.51 & 59.54 & 74.25 & \multicolumn{1}{c}{--}  & 22.94 & 65.87 & 72.33 & 18.97& 76.13 \\
\multicolumn{1}{c|}{} & $\Delta$ASR$_g$ & \multicolumn{1}{c}{--} & 72.30 & 86.44 & 75.54 & 65.00 & \multicolumn{1}{c}{--}  & 78.93 & 85.60 & 91.64 & 91.31 & \multicolumn{1}{c}{--}  & 71.78 & 88.61 & 87.38 & 34.48 & 91.51\\ 
\cline{2-18} 
\multicolumn{1}{c|}{} & CACC & 90.22  & 90.39 & 89.70 & 89.36 & 88.30 & 94.45  & 93.97 & 93.77 & 93.73 & 94.34 & 94.45  & 94.18 & 94.09 & 94.07 & 91.70 & 99.87 \\
\multicolumn{1}{c|}{} & $\Delta$CACC$_p$ & 1.33  & 1.62 & 0.73 & 1.98 & 2.12 & 1.86  & 1.59 & 0.79 & 0.55& 1.86 & 1.87 & 2.07 & 1.61 & 2.59 & 2.43 & 3.32 \\
\multicolumn{1}{c|}{} & $\Delta$CACC$_g$ & 3.01  & 4.65 & 6.37 & 4.36 & 11.97 & 2.81  & 1.67 & 0.79 & 0.55 & 3.04 & 2.81 & 1.53 & 1.61 & 2.59 & 2.07 & 4.77 \\ \bottomrule
\end{tabular}%
}
\caption{Defense performance of PSO and Genetic Algorithm-based defenses. BN denotes BadNet, and RPS denotes RIPPLES.}
\label{tab:my-table}
\end{table*}

\paragraph{Terminate}
We terminate the search process when the global optimization score doesn't increase after one iteration of the update.

\paragraph{Update}
Following previous work, the updating formula of velocity is
\begin{equation}
v^{n}_{d} = \omega v^{n}_{d} +(1-\omega) \times [\Gamma (p^{n}_{d}, x^{n}_{d}) + \Gamma (p^{g}_{d}, x^{n}_{d})]
\end{equation}
where $\omega$ is the inertia weight, and $x_{d}^{n}$, $p_{d}^{n}$, $p_{d}^{g}$ are the d-th dimensions of this particle's current position, individual best position and the global best position respectively. $\Gamma(a,b)$ is defined as 
\begin{equation}
\Gamma(a,b) =\left\{
\begin{aligned}
1 & , & a=b \\
-1 & , & a \neq b
\end{aligned}
\right.
\end{equation}
\par  The initial weight decreases with the increase of numbers of iteration times. The updating formula is
\begin{equation}
\omega = (\omega_{max} - \omega_{min}) \times \frac{T-t}{T} + \omega_{min}
\end{equation}
where $0<\omega_{min} < \omega_{max}<1$, and T and t are the maximum and current number of iteration times.

\par In line with previous work, we update the particle's position in two steps. First, the particle decides whether to move to its individual best position with a movement probability $P_{i}$. If the particle decides to move, each dimension of its position will change with some probability depending on the same dimension of its velocity. Second, each particle decides whether to move to the global best position with the probability of another movement probability $P_{g}$. Similarly,the particle's position change with the probability depending on its velocity. The formulas of updating $P_{i}$ and  $P_{g}$ are 
\begin{equation}
P_{i} = P_{max} - \frac{t}{T} \times (P_{max} - P_{min})
\end{equation}
\begin{equation}
P_{g} = P_{min} - \frac{t}{T} \times (P_{max} - P_{min})
\end{equation}
where $0<P_{min}<P_{max} < 1$.

After mutation, the algorithm returns to the \textbf{Record }step.

\subsection{Genetic Algorithm}
In this section, we will discuss our adapted genetic algorithm (GA) \citep{goldberg1988genetic} in detail following previous notation.

\paragraph{Initialize}
Different from PSO algorithm, we expect the initialized sentences to be more different in order to generate more diverse descendants. So, for each initialization process,  we randomly delete some words and the probability of a word being deleted is randomly chosen among 0.1, 0.2, and 0.3. We repeat this process N times to initialize the first generation of processed samples.

\paragraph{Record}
According to the original GA, we need to compute each individual's fitness in the environment to pick the excellent individuals. Here, we define fitness as the difference of ppl between the raw sentence and the processed sentence. Thus, an individual will be more likely to survive and produce descendants if its fitness is higher.

\paragraph{Terminate}

We terminate the search process when the highest fitness among all individuals doesn't increase after one iteration of the update.

\paragraph{Update}
The update process is divided into two steps. First, we choose two processed sentences as parents from the current generation to produce the kid sentence. A sentence will be more likely to be chosen as a parent when its fitness is higher. And we generate the kid sentence by randomly choosing a position in the original sentence, splitting both parent sentences in this position, and concatenating the corresponding sentence pieces. Second, the generated kid sentence will go through a mutation process. Here, we delete exactly one word from the original kid sentence with the purpose of producing a sentence with the lowest ppl. We repeat this process N times to get the next generation and return to the \textbf{Record} step.

\subsection{Experiments}
\paragraph{Experimental Settings}
For PSO based search algorithm, following previous work, $w_{max}$ and $w_{min}$ are set to 0.8 and 0.2, $P_{max}$ and $P_{min}$ are also set to 0.8 and 0.2. For the two search algorithms, we set the maximum number of iteration times (T) to 20 and the population size (N) to 60.

\paragraph{Results}
Table \ref{tab:my-table} lists the results of two combination optimization based outlier word elimination methods. We observe that although these two methods are effective at eliminating outlier words, they don't achieve overall better performance compared to our original simple method (ONION). Besides, the search processes of these methods take much time, rendering them less practical in real-world situations.

\section{Experiment Running Environment}

For all the experiments, we use a server whose major configurations are as follows: (1) CPU: Intel(R) Xeon(R) E5-2680 v4 @ 2.40GHz, 56 cores; (2) RAM: 125GB; (3) GPU: 8 RTX2080 GPUs, 12GB memory.
The operation system is Ubuntu 18.04.2 LTS (GNU/Linux 4.15.0-108-generic x86\_64).
We use PyTorch\footnote{\url{https://pytorch.org/}} 1.5.0 as the programming framework for the experiments  on neural network models.

\end{document}